\def\year{2021}\relax
\documentclass[letterpaper]{article} % DO NOT CHANGE THIS
\usepackage{aaai21}  % DO NOT CHANGE THIS
\usepackage{times}  % DO NOT CHANGE THIS
\usepackage{helvet} % DO NOT CHANGE THIS
\usepackage{courier}  % DO NOT CHANGE THIS
\usepackage[hyphens]{url}  % DO NOT CHANGE THIS
\usepackage{graphicx} % DO NOT CHANGE THIS
\usepackage{natbib}  % DO NOT CHANGE THIS AND DO NOT ADD ANY OPTIONS TO IT
\usepackage{caption} % DO NOT CHANGE THIS AND DO NOT ADD ANY OPTIONS TO IT
\usepackage{algorithm}
\usepackage{algpseudocode}
\usepackage{algorithmicx}
\usepackage{amsmath}
\usepackage{enumitem}% http://ctan.org/pkg/enumitem
\usepackage{fdsymbol}
\usepackage{bbding}
\usepackage{fontawesome}
\algtext*{EndWhile}% Remove "end while" text
\algtext*{EndIf}% Remove "end if" text
\algtext*{EndFor}% Remove "end for" text

\let\oldReturn\Return
\renewcommand{\Return}{\State\oldReturn}

\algnewcommand\algorithmicto{\textbf{to}}
\algdef{SE}[FOR]{ForTo}{EndFor}[2]{%
  \algorithmicfor\ #1\ \algorithmicto\ #2\ \algorithmicdo
}{%
  \algorithmicend\ \algorithmicfor
}
\algrenewcommand\algorithmicdo{}

\usepackage{color, soul}

\soulregister\cite7
\soulregister\ref7
\soulregister\pageref7

\setstcolor{red}

\title{Commonsense Knowledge Aware Concept Selection For \\ Diverse and Informative Visual Storytelling}
\author{
    Hong Chen\textsuperscript{\rm 1,3}\thanks{With help from the AAAI Publications Committee.}, Yifei Huang\textsuperscript{\rm 1}, Hiroya Takamura\textsuperscript{\rm 2,3}, Hideki Nakayama \textsuperscript{\rm 1}
    \\
}
\affiliations{
    \textsuperscript{\rm 1}The University of Tokyo, 
    \textsuperscript{\rm 2}Tokyo Institute of Technology,\\
    \textsuperscript{\rm 3}National Institute of Advanced Industrial Science and Technology, Japan\\ 
    \{chen, nakayama\}@nlab.ci.i.u-tokyo.ac.jp,       hyf@iis.u-tokyo.ac.jp,     takamura.hiroya@aist.go.jp\\
}
\begin{document}

\maketitle

\begin{abstract}
Visual storytelling is a task of generating relevant and interesting stories for given image sequences. 
In this work we aim at increasing the diversity of the generated stories while preserving the informative content from the images.
We propose to foster the diversity and informativeness of a generated story by using a concept selection module that suggests a set of concept candidates. Then, we utilize a large scale pre-trained model to convert concepts and images into full stories.
To enrich the candidate concepts, a commonsense knowledge graph is created for each image sequence from which the concept candidates are proposed.
To obtain appropriate concepts from the graph, we propose two novel modules that consider the correlation among candidate concepts and the image-concept correlation.
Extensive automatic and human evaluation results demonstrate that our model can produce reasonable concepts. This enables our model to outperform the previous models by a large margin on the diversity and informativeness of the story, while retaining the relevance of the story to the image sequence.
\end{abstract}

\section{Introduction}
% With the recent explosion of multimedia data, the rich and complex information from videos are becoming extremely easy to acquire for a lot of people. When we want to share the video to others, we would describe the story 
Telling a story based on a sequence of images is a natural task for humans and a fundamental problem for machine intelligence for various scenarios such as assisting the visually impaired people. Also known as visual storytelling (VST), the task has raised extensive research attention, since VST requires the model to not only understand the complex content within one image but also reason about the event across images as they occur and change. Since image sequences contain rich and diverse information, it is especially difficult for a model to tell a relevant story that is both informative of the image content and diverse in story style. 

\begin{figure}[t]
\centering
\includegraphics[scale=0.5]{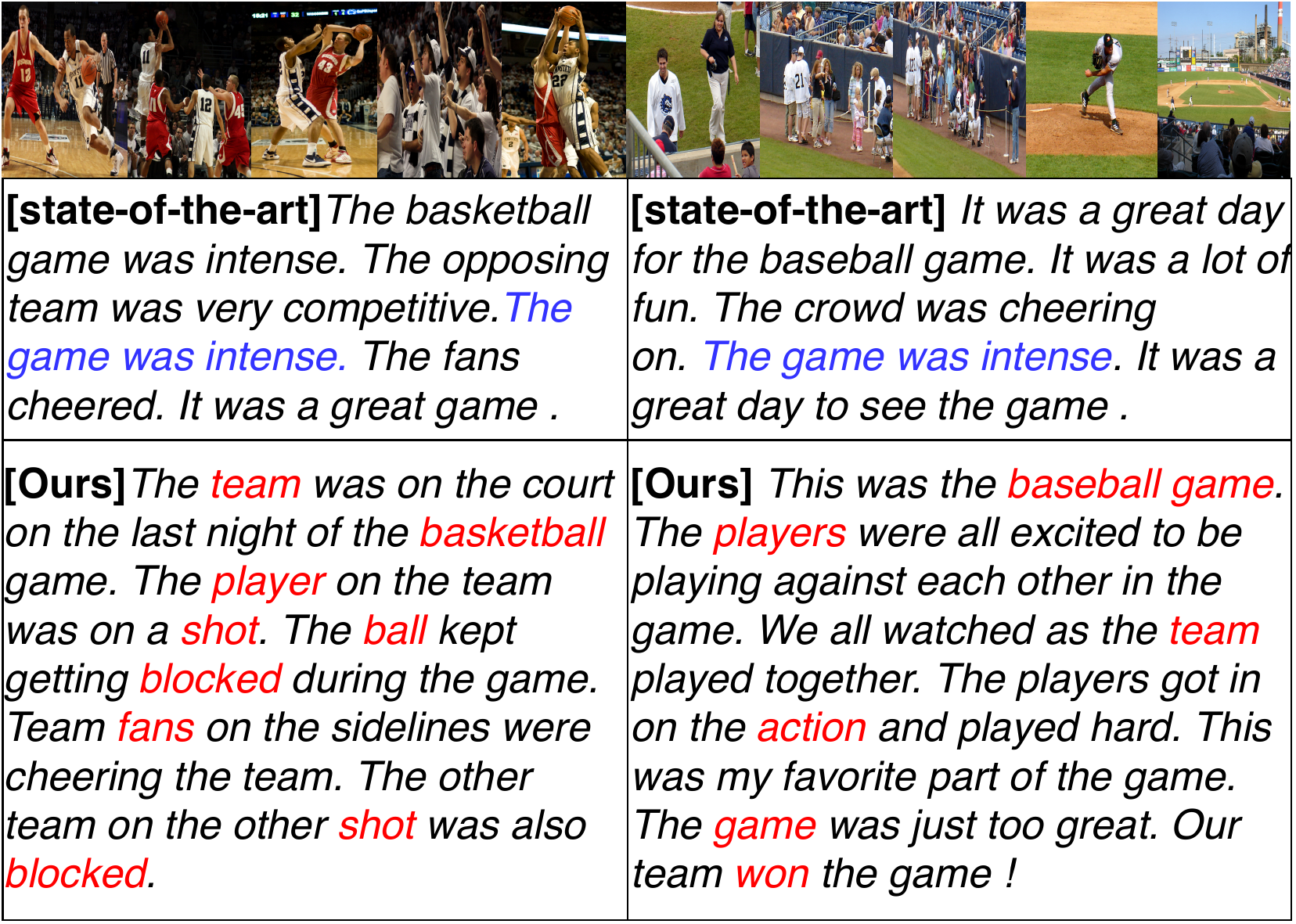}
\caption{Stories generated by the state-of-the-art model~\cite{jung2020hide} and our proposed model using concept selection (red). The state-of-the-art model tends to generate similar stories (blue) for different inputs. Compared with it, our model can generate more informative and diverse stories.}
\label{fig:first}
\end{figure}

Most previous works on VST construct end-to-end frameworks~\cite{yang2019knowledgeable, wang2018no, jung2020hide, yu2017beam}. However, although these methods can produce legitimate stories with high score in automatic metrics like BLEU~\cite{papineni2002bleu}, it is shown that the stories tend to be monotonous which contains limited lexical diversity and knowledge~\cite{hsu2019knowledge} (see the example in Figure~\ref{fig:first}).
% Since the VIST dataset is considerably smaller than other traditional language dataset, end-to-end methods tend to output stories with low diversity and less information~\cite{hsu2019knowledge}.
% \chen{Usually, a visual storytelling model is built under an end-to-end framework of maximum likelihood estimation (MLE) method~\cite{yang2019knowledgeable}. Since the VIST dataset is considerably smaller than other traditional language dataset, end-to-end method tends to output stories with low diversity and less information.}
Recently, two-stage generation methods, also known as plan-write strategy, aroused much research attention in story generation tasks~\cite{yao2019plan,martin2017plan,ammanabrolu2020plan}. When adopted to the task of VST, \citet{hsu2019knowledge} shows that this strategy is capable of generating more diverse stories compared with end-to-end methods. However, their method directly generates concepts from the images using sequence-to-sequence models. Since the concept is selected from the full vocabulary, this kind of direct generation often produces concepts of low quality which affects the informativeness of the story.

\begin{figure*}[h]
\centering
\includegraphics[scale=0.3]{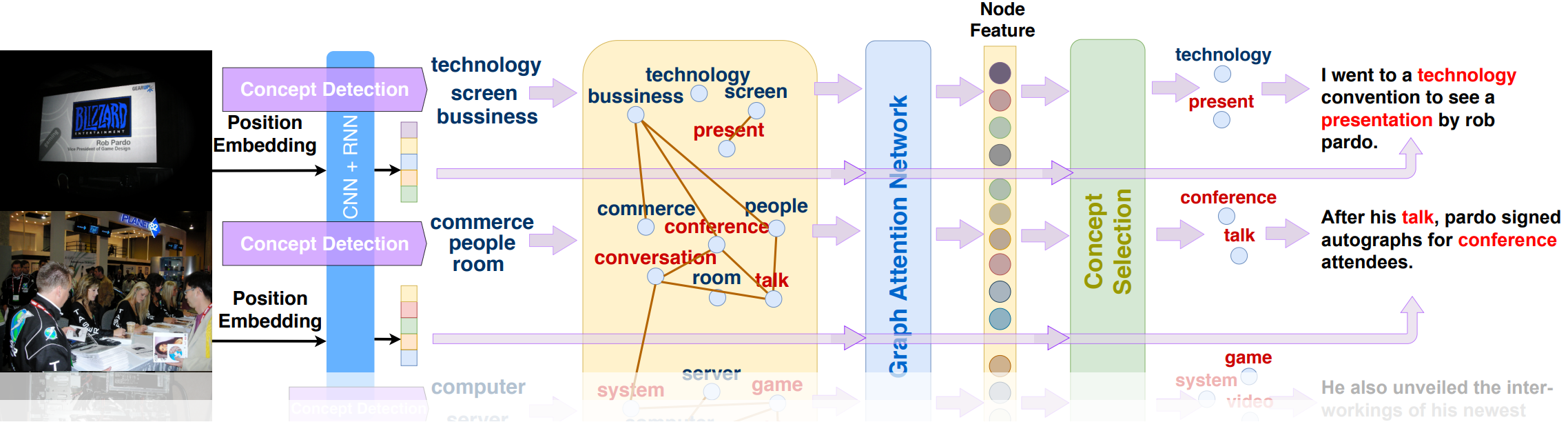}
\caption{An overview of our visual storytelling model. The image features are obtained by a pretrained CNN combined with a bi-LSTM layer. The concepts are obtained from a concept detection model and enriched by ConceptNet~\cite{liu2004conceptnet}. These concepts from the nodes in a graph and are connected according to the relationship in the knowledge base. Initialized by the word embedding vector, the concept features are then updated by a Graph Attention Network. Our proposed concept selection module is then applied to select exact concept words using the image features and the concept features. Finally, both image features and concept words are used to generate a full story.}
\label{fig:overview}

\end{figure*}

In this work we aim to generate stories that are both diverse and informative for a given input image sequence. Taking the advantage of the previous two-stage models, we detect image concepts and construct concept graphs for proposing a set of concept candidates, and propose two novel methods for better selecting the appropriate concept for the second generation stage. After detecting the concept in each input image, we first extend the concepts into a larger commonsense graph using ConceptNet~\cite{liu2004conceptnet}. This extension step increases the informativeness of generated stories. Since selecting appropriate candidates from the concept graph is critical for generating stories of good quality, a natural way is to use a graph attention network~\cite{velivckovic2017graph} to refine the node features. 
Using Graph Attention Network can allow message passing along the graph structure, so that information of related concepts can be updated and integrated. This would allow us to get a better pool of concept candidates.

For selecting the most adequate concept from the candidates as the input to the second stage of our model, two novel modules are proposed in this work. The first one, named Sequential Selection Module (SSM), operates in a straightforward manner that uses an encoder-decoder for selecting concepts for each image. Differently from SSM, the second module called Maximal Clique Selection Module (MCSM) processes the concept graph as a whole. It learns a probability for each concept in the training phase, and during inference it finds a maximal clique using the Bron Kerbosch algorithm \cite{10.1145/362342.362367}. The concepts within the clique are used for the next story generation step. Our experiments show that improved quality of concept selection can greatly help to increase the diversity of the generated stories while keeping the relevance with the input images.

The second stage of our model generates a story with the image features and the selected concepts. Other than using the same module for fair comparison with existing works, we also propose to modify the large scale pre-trained model BART~\cite{lewis2019bart} to input the images and concepts and output the full stories. 

We conduct extensive experiments on the public VIST dataset~\cite{huang2016visual}. Our experiments demonstrate that using our proposed concept selection modules, our generated stories can achieve better performance on both automatic metric and multiple human evaluation metrics using the same generation module. When equipped with BART, the quality of the stories can be remarkably improved, with the generated story diversity similar to human writing.

In summary, our main contributions are listed as follows:
\begin{itemize}[noitemsep]
    \item We propose two novel modules SSM and MCSM to select concepts from the given candidates concepts under a plan-write two-stage visual storytelling system. The experiments show that our proposed methods can output more appropriate concepts than the previous work.
    \item We exploit modified BART as our story generation module to mitigate the problem caused by limited vocabulary and knowledge in the dataset. To the best of our knowledge, this is the first work to use a large scale pre-trained language model in a visual storytelling task. 
    \item Large scale experiments using automatic metrics and human evaluation show that our model can outperform previous models by a large margin in both diversity and informativeness, while retaining the relevance and logicality as the previous work.
\end{itemize}

\begin{figure*}[h]
\centering
\includegraphics[scale=0.4]{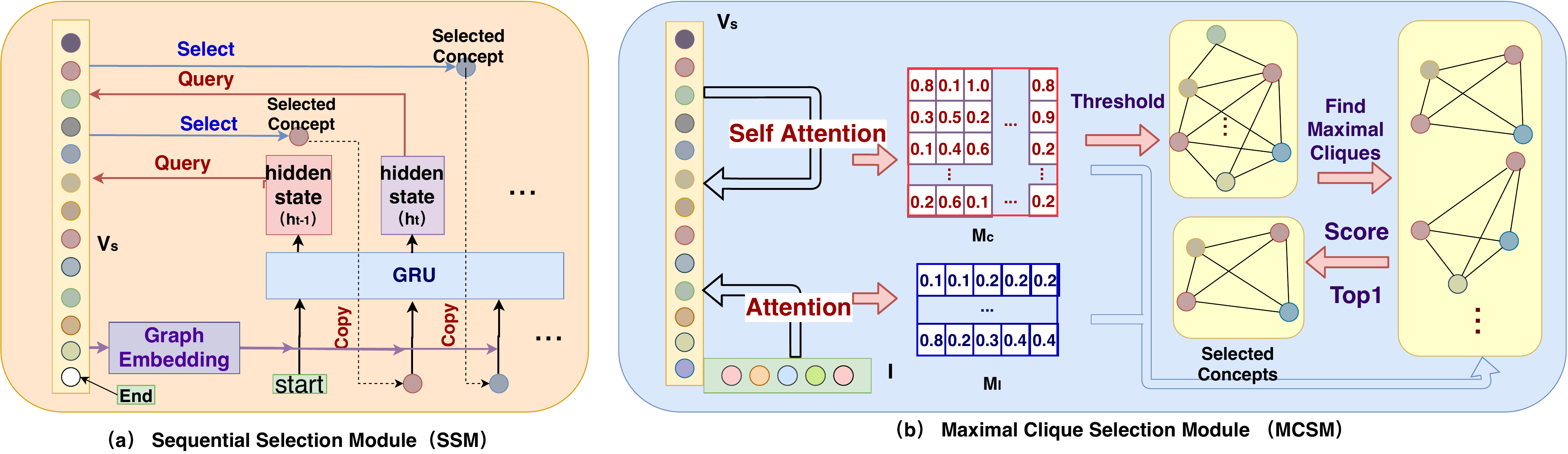}
\caption{
(a) Sequential Selection Module: this module sequentially use current hidden state as a query to select the concept from the commonsense graph.
(b) Maximal Clique Selection Module: this module calculates concept-to-concept and image-to-concept correlation maps. The correlation maps can be viewed as a fully connected graph of concepts. We set a threshold to prune the edges. Then, a maximal clique algorithm is applied to find the maximal cliques in the remaining graph. Finally, those cliques are scored and the one with highest score will be selected and the concepts to be used in the next generation stage.
}

\label{fig:model}
\end{figure*}

\section{Related Work}
Visual storytelling aims at generating stories for image sequences. Many existing methods focused on generating relevant and coherent stories~\cite{hu2020what,story2,story3,story4,story5}. These works can be separated into two lines: one line is to construct end-to-end models to generate the stories directly from the images. The other line is to build a two-stage generation model that first outputs a mid-level abstraction and then generates the full story. 

\subsection{End-to-End Methods}
\citet{wang2018no} proposed a visual storytelling framework which is widely used as a base model in the coming-up studies. This framework uses an end-to-end structure that first convert the image into features and then transfer its information to the adjacent images by a BiLSTM layer. Finally, a decoder decodes the features separately and merge the sentences into a story.
While many succeeding works \cite{huang2019hierarchically,jung2020hide} can achieve high automatic scores, the story may not be interesting and informative~\cite{hu2020what} for human as they often contain repetitive texts and limited information.
One of the main reasons for the low diversity and informativeness is that these model are trained end-to-end under the maximum likelihood estimation~\cite{yang2019knowledgeable}.

\subsection{Two-stage Methods}
To alleviate the low diversity problem, \citet{hsu2019knowledge} proposed to generate several concepts before outputting the full stories. The discrete concept words can guide the decoder to produce more diverse stories. This plan-and-write strategy~\cite{yao2019plan} can substantially increase the diversity of the stories. 
During the planning, to enhance the concepts that models can obtain, some researchers~\cite{hsu2019knowledge,yang2019knowledgeable} introduce external commonsense knowledge database such as OpenIE~\cite{angeli2015openie}, Visual Genome~\cite{krishna2017genome} or ConceptNet in the VST task. Their results show that using external knowledge base helps to generate more informative sentences. These works also show that the concepts are critical for the quality of generated stories because they can control the story flow.

In this work we aim to improve the concept selection for increasing the diversity and informativeness of VST. We propose two concept selection modules that carefully selects concept from a well-designed pool of concept candidates. The stories generated using our selected concepts thus become more diverse and informative. We further introduce to modify the pretrained model BART and use it to generate even better stories.
% Different from the existing works, in this paper planning model and generation model are separated trained because this can helps us to better control the story flow and become easy to assess the accuracy for either parts. The planned works directly impact the story generated in the second step. So we propose two novel concept selection module to select more appropriate concepts.

% Meanwhile, incorporating different decoding methods is also an effective way to improve diversity. For example, \citet{hsu2018using} propose a inter-story diverse beam search to reduce the repetition in different stories. 
% In this paper, we apply nucleus sampling~\cite{holtzman2019curious}, a recent proposed decoding method that shows great improvement on story generation tasks.
% \section{Background:Increasing Diversity}

% In this paper, we introduce two ways to increase diversity.

% \subsubsection{planning before generation}
% As mentioned before, planning explicit words as the backbone of the stories can increase diversity which was proved in other story generation tasks \cite{yao2019plan}. The generated stories can be well controlled by these words. More appropriate planning can produce stories with better performance.

\section{Method}
% Our model is composed of (在这里介绍一下overall, 不一定说our model is composed of, 也可以说Given Image I as input, ... 然后说在后面的subsection中我们会依次介绍)
% 按照data flow的顺序介绍model, [commonsense extraction, image feature encoding],  [graph construction, GAT], [concept selection], [最后的bart], 可以是一个中括号对应一个subsection.

Figure~\ref{fig:overview} depicts an overview of our proposed model.
Given a sequence of $N$ images as input, our model 1) encode image features, 2) construct a large commonsense graph for the images, 3) update concept feature in the graph, 4) select the concepts from the graph and 5), send concepts and image features into the decoder to output the full story. The details of each step are as follows:

\subsection{Image Feature Encoding}
We send the images into ResNet152~\cite{7780459} to obtain image features $I = \{I_1, ..., I_N\}$. Following \citet{wang2018no}, a bidirectional GRU further encodes the high-level visual features to capture sequential information. The encoded feature for each image contains both the information of itself and the information from the adjacent images. Note that position embedding is applied before sending the image features into GRU to identify the order of the images.

\subsection{Commonsense Graph Construction}
To build our commonsense knowledge graph for image sequences, we need some seed concepts.
Following \citet{yang2019knowledgeable}, we use clarifai~\cite{clarifai} to obtain the top 10 seed concepts from each image.
Each concept is used as a query to select relative commonsense concepts in the ConceptNet~\cite{liu2004conceptnet}.
Since the number of the commonsense concepts is usually very large $(>500)$, we make several rules to filter some concepts which are less useful:
\begin{itemize}[noitemsep]
    \item Remove the commonsense concepts that appear less than 5 times in the whole training corpus.
    \item Remove the commonsense concepts that do not co-occur with the seed concepts either in the sentence or in the training corpus.
    \item If the concept number is still larger than $K$ for one image. We simply randomly sample $K$ words from it.
\end{itemize}
After the filtering process, each image contains $K$ concepts. Note that while different images can obtain the same concept in one image sequence, they can represent different semantic meanings in different positions of the story.
% connect the concepts
Each concept forms a node in the commonsense graph. An undirected edge is established between concepts if they are related in ConceptNet. Also, a concept in one image will connect to the related concepts in the adjacent images to allow information flow between images. Like~\cite{yang2019knowledgeable}, we do not use the specific relation (\textit{e.g.}, isA, has) between concepts.
Till now, we build a graph which is a graph structure that is both connected within an image and between images.

\subsection{Concept Features Update}
The concept features are initialized with word embedding vectors. 
To incorporate the visual information into the concepts, we also connect the image feature to its corresponding concept features in the graph.
These features are updated by a two-layer Graph Attention Network, which passes information between connected concepts and image using attention mechanism.

\subsection{Concept Selection Module}
We propose two methods to select concepts given the concept features and the image features.

To better formalize the procedure in the methods,
we denote $c^{i,j}$ as the j-th concept of the i-th ($1 \leq i \leq N$) image.
we let $\mathcal{C}_S=\{c_S^{1,1}, ..., c_S^{N,K}\}$ and $\mathcal{C}_G = \{c_G^{1,1}, ...\}$ denote the concepts set in the source candidate concepts and the full word set in the gold story, respectively. The target concepts are their intersection: $\mathcal{C}_T = \mathcal{C}_S \cap \mathcal{C}_G$.
% Each set can be described as $\bigcup_{i=1}^N \mathcal{C}_*^N$ which is a combination of the concept or word set of $N$-th image.

\subsubsection{Sequential Selection Module (SSM)}
One straightforward way of selecting concepts is to adopt an encoder-decoder model where we can forward the updated concept features into the encoder, and the decoder will output the selected concepts.
Inspired by the Copy Mechanism~\cite{gu201copy}, instead of generating a probability distribution with vocabulary size in each step, the SSM outputs are directly chosen from the inputs $\mathcal{C}_S$. 
% As shown in Figure~\ref{fig:model}(a), instead of projecting the hidden state into a probability distribution with vocabulary size, we take the t-1 step hidden state $h^{t-1} \in (1\times D)$ and concept feature $v_S^{t-1} \in (1\times D)$ selected in $t-1$ step as a query to predict the score $s$ for all candidate concept features $V_S\in (NK\times D)$. Here, we use two trainable weights $W_{h}$ and $W_{c}$ as projection matrix for hidden state and candidate concept features. After that, a softmax activation function is applied for us to select the concepts $c_S^t$ with the highest score.
As shown in Figure~\ref{fig:model}(a), we use a GRU~\cite{cho2014learning} to first encode the concept embedding feature $v_S^{t-1}$ and the hidden state into a new hidden state $h^t$. We then use $h^t$ to query all the concepts in $\mathcal{C}_S$ to get a probability $p_S$ for each concept in the source set. Finally the concept with the highest probability is selected as the output concept, while its feature is directly copied for the generation of the next step:
\begin{align}
\begin{split}
h^{t} &=GRU\left(h^{t-1},v^{t-1}_S \right) \\
p_S &= softmax\left((W_{h} h^{t})^\mathrm{T} W_{c} V_S\right) \\
\label{equ:selection}
c_S^t &= argmax(p_S)
\end{split}
\end{align}
Here $W_{h}$ and $W_{c}$ are trainable projection matrices.
The objective function is to maximize the probability score of the concepts which locate in $C_T$.
\begin{align}
\mathcal{L}_{ssm} &= -\Sigma y_{S,T}\log (p_S),
\end{align}
where $y_{S,T}$ is an indicator of whether a concept in $\mathcal{C}_S$ is in $\mathcal{C}_T$.
The sequence selection step stops when the module generates $<$end$>$ token. This $<$end$>$ token is added to the set of candidate concepts with a uniform random initialized feature without any update during the training phase. The same procedure is done to the $<$start$>$ token except that it is not involved in the candidates.

\subsubsection{Maximal Clique Selection Module (MCSM)}

Different from SSM, this method aims to calculate the co-occurrence probability of all candidate concepts $c_s$ in the graph. An illustration of MCSM is shown in Figure~\ref{fig:model}(b). In the beginning, we calculate self-attention to compute a correlation matrix $M_C \in (NK\times NK)$ which contains the correlation between each pair of nodes. We also calculate another correlation matrix for each image $M_I \in (N\times K)$ indicating the correlation between the concept embedding feature ($v_S$) and image features ($I$).
\begin{align}
\begin{split}
    M_C &= \sigma(v_S^\mathrm{T} W_a^\mathrm{T} W_b v_S) \\
    M_I &= \sigma(I^\mathrm{T} W_c^\mathrm{T} W_d v_S)
\end{split}
\end{align}
Here, $W_a, W_b, W_c, W_d$ is trainable weights, $\sigma$ denotes sigmoid activation function.
Intuitively, the concepts that appear in a gold story should own high correlations with each other, and the image should be highly related to the gold concepts to describe it. Thus, our target correlation maps can be written as follow:
\begin{align}
\begin{split}
    \hat{M}_C[i,j] &= 
    \begin{cases}
    1,             &c_i\in \mathcal{C}_T \wedge c_j \in \mathcal{C}_T\\
    0,                          &\text{otherwise}    \\
    \end{cases}\\
    \hat{M}_I[i,j] &= 
    \begin{cases}
    1,             &c_j \in \mathcal{C}^i_T \\
    0,             &\text{otherwise}    \\
    \end{cases}
\end{split}
\end{align}
Then, the objective is to minimize the difference between predicted and target correlation maps:
\begin{align}
    \mathcal{L}_{mcsm} = \lambda_1\left \| M_C-\hat{M}_C \right\|_2^2 + \lambda_2\left \| M_I-\hat{M}_I \right\|_2^2
\end{align}

% To select the concepts as the inference to generate story,
In testing phase, $M_C$ can be viewed as a fully connected graph in which the edge weights correspond to the values in the matrix.
Therefore, a low edge weight means less co-occurrence probability between two concepts. Based on this assumption, we set a threshold $\tau$ to remove the edges whose weight is less than $\tau$. Then we apply Bron Kerbosch algorithm \cite{10.1145/362342.362367} to find all maximal cliques from the remaining sub-graph. 
% The minimum clique length is set to 7 in our experiment. 
Finally, we score each of them with equation~\ref{equ:score} and select a clique with maximum score $s$. The output concepts are the nodes within the selected cliques.

\begin{align}
\begin{split}
    s &= s_C + s_I\\
    s_C &= \frac{1}{(\left \| \mathcal{C}_P  \right\|-1)\left \|\mathcal{C}_P\right\|}\sum_{i}\sum_{j\neq i}log(M_C[i,j])\\
    s_I &= \frac{1}{\left \| \mathcal{C}_P  \right\|}\sum_{i=1}^N\sum_{c_j \in \mathcal{C}_P^i}log(M_I[i,j]).
    \label{equ:score}
\end{split}
\end{align}
where $\mathcal{C}_P$ denotes the concepts in a clique, and $\mathcal{C}_P^i$ denotes the concept of the i-th image in the clique.

%need a algorithm
% \begin{algorithm}[t]
% \begin{algorithmic}[1]
%     \Function{MCSM}{$C$, $I$}
%         \State{$IC_{|I|*|C|} = Attention(I, C)$;}
%         \State{$CC_{|C|*|C|} = Attention(C, C)$;}
        
%         \State{$\tilde{CC} = Pruning(CC, t)$}
%         \State{$Cliques = Born\_Kerbosch(\tilde{CC}, min_len)$}        
%         \State{$Clique_{score} = Score(Cliques)$}        
%         \Return{$Clique_{score}.top1()$}

%     \EndFunction
% \end{algorithmic}
% \caption{Maximal Clique Selection Module (MCSM)}
% \label{alg:planning}
% \end{algorithm}

\subsection{Concept to Story Module}
The selected concepts are assigned to its corresponding image to generate the sentences.
We tried two kinds of encoder-decoder to decode the story:
1) a simple encoder-decoder module that uses multi-head pooling to encode the concept embeddings, and decode the sentences with a RNN decoder.
2) a large scale encoder-decoder which both can encode the input and output the sentences.

RNN Decoder decodes sentences separately for each image and then concatenates into a story, while BART accepts all the images and concepts at once to output a full story.

\subsubsection{RNN}
To get the concept feature from the concept words, we apply Multi-head Pooling~\cite{liu2019pool} to calculate multi head self-attention score on the input concept embedding and do weighted summation on the concept embedding. 
Each image and corresponding concepts are decoded separately with the same decoder as \citet{jung2020hide}.
The decoder accepts the image features $I_i$ and the pooled concept feature $v_i$ as input.
Formally, the generation process can be written as:
\begin{align}
h^{t}_{i} &=RNN\left(h^{t-1}_{i},\left[w_{t-1}^{i}; I_i; v_i \right]\right) \\
\pi^{t} &= softmax\left(W_{s} h^{t}_{i}\right)
\end{align}
% Here $h^{t}_{i}$ denotes the hidden state in step $t$ of image $i$ and is calculated by the previous hidden state $h^{t-1}_{i}$ and a concatenation vector of previous output token embedding $w^{t-1}_{i}$, image feature $I_i$ and concept feature $c_i$.
Here $W_s$ is a projection matrix for outputting the probability distribution $\pi^{t}$ on the full vocabulary. ``;" denotes the channel-wise concatenation operation. Finally, all of the corresponding words are merged into a complete story.

\subsubsection{BART}
Although there exists many large scale pretrained language models, none of them are used on VST in previous works. In this paper, we propose to use a modified version of BART as our base decoder model. BART is pretrained on a large dataset and their arbitrary noise function helps to mitigate the exposure bias.
% \begin{itemize}[noitemsep]
%     \item Book corpus Dataset is in the pre-trained dataset which is a large dataset of story.
%     \item Their arbitrary noise function applied in token may help mitigating the exposure bias.
%     \item It is easily modified to concept-to-story generation that requires both concepts and images as input.
% \end{itemize}
For inputting the image features into BART, We add a new randomly initialized projection matrix to transform the size of image features to fit BART's input. 
For the concepts we simply use the pretrained BART word embedding.
Since BART is powerful enough to generate a long story. Different with RNN decoder, we input N images and all concepts into the BART together and use a $<$\textit{SEP}$>$ token to separate the images and concepts, and to separate the concepts from different images.
The transformed image feature and concept embedding are sent to BART and generate the final stories. 
% 下面这一句可以写在detail或者supplementary里面
% The finetuning contains two steps: 1) freeze all the parameters except for the image projection matrix. 2) train all model parameters for a small number of iterations.

% \begin{figure}[t]
% \centering
% \includegraphics[scale=0.25]{LaTeX/figure/bart.pdf}
% \caption{blablabla}
% \label{fig:overview}
% \end{figure}

% \begin{figure}[t]
% \centering
% \includegraphics[scale=0.45]{LaTeX/figure/rank.pdf}
% \caption{The y axis shows the transformation of Ranking score(lower the better) and Distinct-4 score(larger the better) while we change the temperature in nucleus sampling with the model trained from scratch on VIST dataset. 
% We ask the workers to rank the overall score for five stories generated by 5 different temperatures.
% As we can see, with the increasing of the temperature, the stories become more diverse, however, the quality of them become lower.}
% \label{fig:rank}
% \end{figure}

\section{Experiment}
\subsection{Dataset}

We conduct experiments on the widely used VIST dataset \cite{huang2016visual}, which consists of 40101 samples for training 5050 test samples. Each sample contains 5 images and one corresponding gold story. 
For fair comparison, we follow the same experiment setting as \cite{jung2020hide} except that we set the vocabulary size to 28000.
All models use the same fixed random seed.

\subsection{Concept Detector}
In this paper, we use the Concept Detection API produced by clarifai \footnote{\label{footnote-1}www.clarifai.com}. This Detector can predict 11,000 concepts which is larger than any other detection model. This powerful pretrained detector helps us to precisely find out the concepts inside the images, so that the word in the gold sentences can be easily involved in our knowledge enhanced candidate concepts.

\subsection{Implementation Details}

When training SSM, since we assume that there is no order relationship between the concepts in one image, so during the training phase, we randomly shuffle the target concept in one image.
When training BART, we conduct a two-stage training: 1) freeze all BART parameters except for the image projection matrix. 2) finetune all parameters. We use Adam~\cite{kingma2014adam} optimizer with an initial learning rate of 4e-4 in the first stage, then the learning rate is decreased to 4e-5 in the fine tuning stage. Each stage is trained for 5 epochs.
All the other parts of our model are trained with an Adam Optimizer with learning rate 4e-4.
During training, we follow \citet{jung2020hide} to blind one of the images starting from the 50-th epoch and increase the blinding into two images from epoch 80. The training stops at epoch 100.
Our model uses gold concepts extracted from gold stories to train the concept to story model. This step is similar to the common auto-regressive models that use the target token as the input to generate the next token. As has been discussed in the previous sections, this kind of generation often meets with the problem caused by the train-test discrepancy that we cannot see the gold concepts in the testing phase. To alleviate, a simple and effective way is to add noise to the inputs. In this work, we add two kinds of noise into the inputs in the story generation module: masking and random replacement. We mask 30\% concepts and replace 20\% of them into other similar words in training.

\subsection{$\tau$ search in Maximal Clique Selection Module}
 $\tau$ is set as the threshold in pruning edges for the maximal clique selection. Larger $\tau$ leads to fewer concepts that can be selected and will further result in the lack of imagination (diversity) in the generated story. However, smaller $\tau$ would lead to too many concepts that may mislead the model to generate irrelevant stories. To make a trade-off, we initialize $\tau$ as 0.3 and continual decreasing the number until Bron Kerbosch algorithm \cite{10.1145/362342.362367} can produce at least 5 candidate cliques that contains 7 to 15 candidate concepts in each clique.

\begin{table}[t]
\centering
\begin{tabular}{l|c|c|c}
\hline
 Method&  Precision& Recall&  F measure \\\hline\hline
 Rand&  2.68&  2.45&  2.56 \\\hline
 C\_Attn&  30.38&  \textbf{43.37}&  35.86 \\\hline
 I2C&  31.32&  20.75&  24.96\\\hline
 SSM&  40.43&  40.30&  40.36\\\hline
 MCSM&  \textbf{45.30}&  40.90&  \textbf{42.99}\\
  \hline
\end{tabular}
\caption{Concept selection performance of different methods. The results show that our MCSM achieved the best f-score among all methods.}
\label{tab:acc}
\end{table}

\subsection{Decoding strategy}

\begin{figure}[t]
\centering
\includegraphics[scale=0.45]{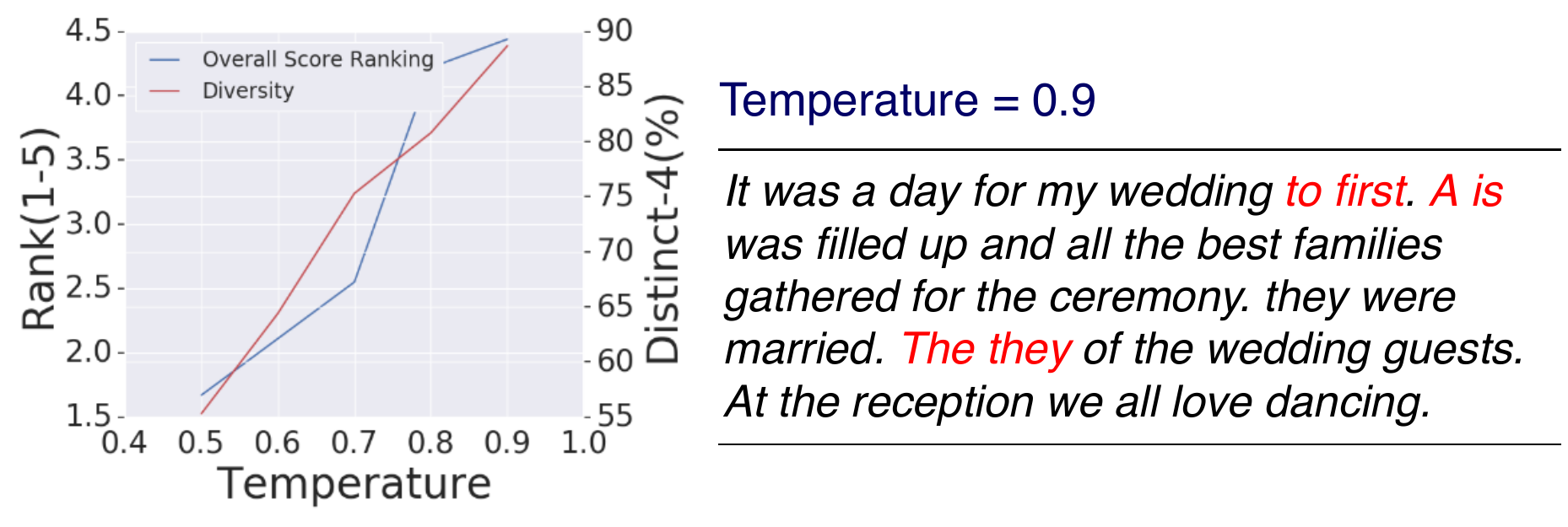}
\caption{The y axis shows the ranking score (lower the better) and Distinct-4 score (higher the better) while we change the temperature in nucleus sampling with the model trained from scratch on VIST dataset. 
We ask the workers to rank the overall score for five stories generated by 5 different temperatures.
As we can see, with the increasing of the temperature, the stories become more diverse, however, the quality of them become lower.}
\label{fig:rank}
\end{figure}

During model inference, usually beam search is used in decoding the sentences from the decoder~\cite{wang2018no, jung2020hide, yu2017beam}. However, it has been proved that beam search can result in bland, incoherent stories, or gets stuck in repetitive outputs~\cite{holtzman2019curious}. In this paper, to further improve diversity in generated stories, we apply nucleus sampling~\cite{holtzman2019curious} which outperforms top-k sampling method. However, nucleus sampling does not perform well without a well-trained language model. We show this by an experiment where we train our model with RNN as the story generation module on the VIST dataset and decode using nucleus sampling. Note the VIST dataset is relatively small compared with other natural language datasets like Book Corpus~\cite{Zhu_2015_ICCV}. We conduct human evaluation and let the annotators to rank the overall story qualities (lower the better) generated with different softmax temperatures [0.5,0.6,0.7,0.8,0.9] for the same image sequence. We randomly pick 100 samples for this experiment. Figure~\ref{fig:rank} shows that with the temperature increases, the story quality would drop.
Especially, when the temperature is 0.9 which is recommended in \citet{holtzman2019curious}, the generated sentences are incomprehensible and full of grammatical errors, as shown in the example of Figure~\ref{fig:rank}. This result motivates us to use a large scale pre-trained language model in our experiment.

Thus, for fair comparison with previous works, the same RNN with beam search is used as the story generation module to validate the effectiveness of our concept planning model.
While, we use nucleus sampling (temperature=0.9, p=0.9) when BART \cite{lewis2019bart} is applied to demonstrate a higher upper bound of our model's ability.

\subsection{Experiment on Concept Selection}
We first test the ability to select appropriate concepts for different models.
Since each image sequence in the test sample corresponds to 3 to 5 gold stories, we report the highest performance of the output selection result with respect to all gold stories in each sample. 
Similar as Keyphrase generation tasks, we apply precision, recall and f measure to evaluate the efficiency of concept selection methods. The precision is defined as the number of correct concepts selected over the target concepts, and recall is defined as the number of correct selections over all candidates.

We compare among several methods:
\begin{itemize}[noitemsep]
    \item \textbf{Rand}: A simple baseline where we randomly pick 3 concepts from the candidates for each image. On average, each image contains 2.65 gold concepts.
    \item \textbf{C\_Attn}: We extract the attended concepts where the attention score is larger than a threshold from the model of \citet{yang2019knowledgeable}. This is an end-to-end model with sigmoid attention on concept words. We choose 0.8 as the threshold since this contributes the best f-score.
    \item Image to concept(\textbf{I2C}): This is a straightforward version of concept selection where the concepts are directly generated from the images. We simply add a projection layer on each hidden state to predict the concept words from the vocabulary size of the concepts, which is very similar to the model of \citet{hsu2019knowledge}.
    \item \textbf{SSM}: Our proposal which uses a copy mechanism in each step of selection.
    \item \textbf{MCSM}: Our proposal which calculates the correlation score for concept-concept and image-concept and uses maximal clique selection.
\end{itemize}

\begin{table}[t]
\centering
\begin{tabular}{l|c|c|c}
\hline
 Method& Dist-2& Dist-3 &Dist-4\\\hline\hline
 INet$\smwhitestar$ &8.36 &18.92 &31.02\\\hline
 KS$\smwhitestar$ &10.84 &22.90 &36.51\\\hline
 KG-Story\dag &18.73 &38.65 & 57.22\\ \hline
 Our(MCSM) &13.98 & 34.01& 54.11\\ \hline
 Image+BART\dag &21.63 & 46.23& 67.57\\ \hline
 Our(MCSM)+BART\dag & \textbf{34.95}& \textbf{69.88}& \textbf{88.74}\\ \hline
 \hline
 Gold                 & 47.76& 82.27 & 95.05\\ \hline
\end{tabular}
\caption{Diversity of generated stories by different methods. Two-stage generation methods can produce more diverse stories. Using BART, we can achieve the diversity close the human writing, while achieving same level story quality in other aspect. \dag denotes the story generation module is pre-trained with other dataset. $\smwhitestar$ denotes end-to-end methods.}
\label{tab:diverse}
\end{table}

Qualitative results are shown in Table~\ref{tab:acc}. We can see that our proposed SSM and MCSM can achieve significantly higher f-score than other methods. This helps our model to keep the story relevance to the input images while generating diverse stories.

\subsection{Experiment on Visual Storytelling}
Here we show the results of the visual storytelling. We use the following baselines for comparison:

\noindent\textbf{INet}~\cite{jung2020hide} This is a recent work which uses a ``hide-and-tell" strategy to train an end-to-end model. In this method no concept is used.

\noindent\textbf{{KS}}~\cite{yang2019knowledgeable} This method uses sigmoid attention to incorporate concept features into the model. We change the structure of the visual encoder and decoder the same as \textbf{INet} for fair comparison.

\noindent\textbf{KG-Story}\dag~\cite{hsu2019knowledge} is a strong baseline that use two stage plan-write strategy and pretrain the decoder on RocStories Corpora~\cite{mostafazadeh2017lsdsem}. \dag indicates the model introduces external large-scale pretraining data.

\noindent\textbf{Image+BART}\dag is an end-to-end baseline that uses BART on top of image features to directly generate the story. This baseline is one-stage that does not generate concepts.

We also change the concept selection module and story generation module in our model to validate the effectiveness of each component. Specifically, we compare: \textbf{Rand+RNN}, \textbf{C\_Attn+RNN}, \textbf{SSM+RNN}, \textbf{MCSM+RNN}, and \textbf{MCSM+BART}\dag~.

\subsubsection{Comparison on diversity}
We first compare the ability of generating diverse stories of different models. Quantitative comparison is shown in Table \ref{tab:diverse}. We report Distinct-n (Dist-n) scores~\cite{li2015dist} that calculate the percentage of unique n-gram in all generated stories in the test dataset. Higher score means less inter-story repetition. 
From the table, two stage generation methods (KG-Story and ours) can achieve significantly higher diversity scores. 
Our MCSM can generate the most diverse stories among all the methods without using external pretrained models.
When equipped with BART, we can even achieve diversity close to human writing.
We show in the following that the increased diversity also improves the overall quality of the story.

\begin{table}[t]
\centering
\begin{tabular}{l|c|c|c|c|c}
\hline
 Method& B-3& B-4 & R& M & C\\\hline\hline
 INet$\smwhitestar$& 23.5&14.4&29.7&35.3&9.5\\\hline
 KS$\smwhitestar$&\textbf{24.7}&\textbf{15.0}&\textbf{31.0}&35.0&9.8\\\hline
 \hline
 Rand+RNN&  13.3&6.1&27.2&31.1&2.2\\\hline
 C\_Attn+RNN&20.7&11.2&29.7&34.5&7.8\\\hline
 SSM+RNN &22.1&12.0&30.0&35.4&10.5\\\hline
 MCSM+RNN&23.1&13.0&30.7&\textbf{36.1}&\textbf{11.0}\\
 \hline
\end{tabular}
\caption{Automatic metric in story quality. We report BLEU (B), METEOR (M), ROUGH-L (R), and CIDEr (C) scores. The two-stage generation can achieve higher METEOR and CIDER scores.}
\label{tab:am}
\end{table}

\begin{table*}[!ht]
\footnotesize
\centering
\small
\begin{tabular}{l||c|c||c|c||c|c||c|c||c|c}
    \hline
    Choices(\%) & \multicolumn{2}{c||}{MCSM vs INet}& \multicolumn{2}{c||}{MCSM vs KS}& \multicolumn{2}{c||}{MCSM vs SSM}& \multicolumn{2}{c||}{MCSM+BART\dag ~vs KS}& \multicolumn{2}{c}{MCSM+BART\dag ~vs MCSM}\\\cline{2-11}
    & MCSM & INet & MCSM & KS & MCSM & SSM & MCSM+BART & KS & MCSM+BART & MCSM\\
    \hline
    Revelence               &\textbf{47.4} &35.6    &26.3&\textbf{31.6}     &\textbf{50.5} &40.0                 &28.8 &\textbf{33.6}                        &35.2&35.2\\
    Informativeness         &\textbf{51.0*} &31.6    &\textbf{46.3*} &28.9    &\textbf{44.7} &41.2                 &\textbf{62.5**} &18.8                       &\textbf{58.8**} &23.5\\
    Logicality                 &\textbf{35.5} &34.3    &\textbf{34.2} &29.0    &32.9 &\textbf{42.3}                 &\textbf{35.3} &33.3                        &\textbf{40.2} & 37.5\\
    Overall                 &\textbf{55.0**} &30.0    &\textbf{44.7} &34.2    &\textbf{48.3} &37.1                 &\textbf{43.5**} &23.0                        &\textbf{47.0*} & 31.6\\
    
    \hline
\end{tabular}
\caption{Human evaluation. Numbers indicate the percentage of annotators believe that a model outperforms its opponent. Methods without (+BART) means using RNN as the story generation module. Cohen’s Kappa coefficients ($\kappa$) for all evaluations are in Moderate (0.4-0.6) or Fair (0.2-0.4) agreement, which ensures inter-annotator agreement. We also conduct a sign test to check the significance of the differences. The scores marked with * denotes $p < 0.05$ and ** indicates $p < 0.01$ in sign test.}

\label{tab:human}
\end{table*}

\subsubsection{Automatic evaluation}
In Table~\ref{tab:am}, for comparing the quality of generated stories, we use automatic metrics BLEU (B)~\cite{papineni2002bleu}, METEOR (M)~\cite{banerjee2005meteor},
ROUGH-L (R)~\cite{lin2004rouge}, and CIDEr (C)~\cite{vedantam2015cider}. Note that it remains tricky for automatic scores to appropriately evaluate story qualities. Using concept, KS can achieve better performance than INet that does not use concept. From the comparison of the variants of our model, we can see that better concept selection can lead to better automatic scores. With reasonable concept selection, our SSM and MCSM can achieve highest METEOR and CIDER scores.

\subsubsection{Human Evaluation}
To better evaluate the quality of generated stories, we conduct human evaluation to compare pair-wise outputs with several models via the Amazon Mechanical Turk (AMT). 
We randomly sample 200 image sequences from the test set and generate stories using each model. For each sample pair, two annotators participate in the judgement and decide their preference on either story (or tie) in terms of Relevance, Informativeness, Logicality and Overall. \textbf{Relevance} evaluates how relevant the stories and the images are. \textbf{Informativeness} assesses how much information can be achieved in the generated stories, and this score from one side reflects the diversity of stories. \textbf{Logicality} evaluates the logical coherence in the stories. \textbf{Overall} is a subjective criterion that shows the preference of workers.

Table~\ref{tab:human} shows the human evaluation result. Since there exists randomness in human evaluation, we compute the Cohen's Kappa coefficient and found that all evaluations are in Moderate Agreement and Fair agreement, which indicates the evaluation result is reasonable as good inner agreement between evaluators is reached. We also conduct a Sign test to illustrate the significance of the evaluation difference: if p is below 0.05 it would indicate that the two compared models have a significant performance difference.
From the comparison between MCSM and INet and the comparison between MCSM and KS, we can see that our two-stage planning method greatly outperforms the end-to-end models, especially in the informativeness score. The MCSM module also outperforms the SSM module, which indicates positive correlation between the quality of concept selection and the overall quality of generated stories.
Finally, using BART with MCSM can help to achieve further informativeness and generate even better stories.

\begin{figure}[t]
\centering
\includegraphics[scale=0.3]{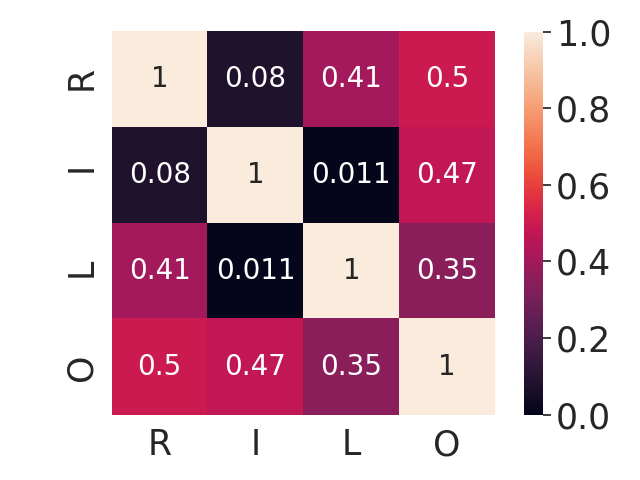}
\caption{We calculate the Pearson Correlation Coefficient for each criteria in human evaluation. R, I, L, O denotes Relevance, Informativeness, Logicality and Overall, respectively. We can see that Informativeness is almost independent to Relevance and Logicality, while is highly correlated to Overall score.}
\label{fig:correlation}
\end{figure}

\subsection{Importance of informativeness in story quality}
We calculate the Pearson Correlation Coefficient on four criteria in human evaluation. 
In Figure~\ref{fig:correlation}, R,  I,  L,  O  denotes Relevance, Informativeness, Logicality and Overall, respectively. 
Ranged from -1 to 1, the Informativeness score has low correlation score with Logicality (0.011) and Relevance (0.08), while a high correlation score with Overall (0.47).
This indicates that Informativeness is almost independent on Relevance and Logicality, but highly dependent on the Overall score. This suggests that humans tend to choose stories with more interesting information. This phenomenon proves the significance of informativeness and diversity in visual storytelling.

\begin{figure}[!t]
\centering
\includegraphics[scale=0.33]{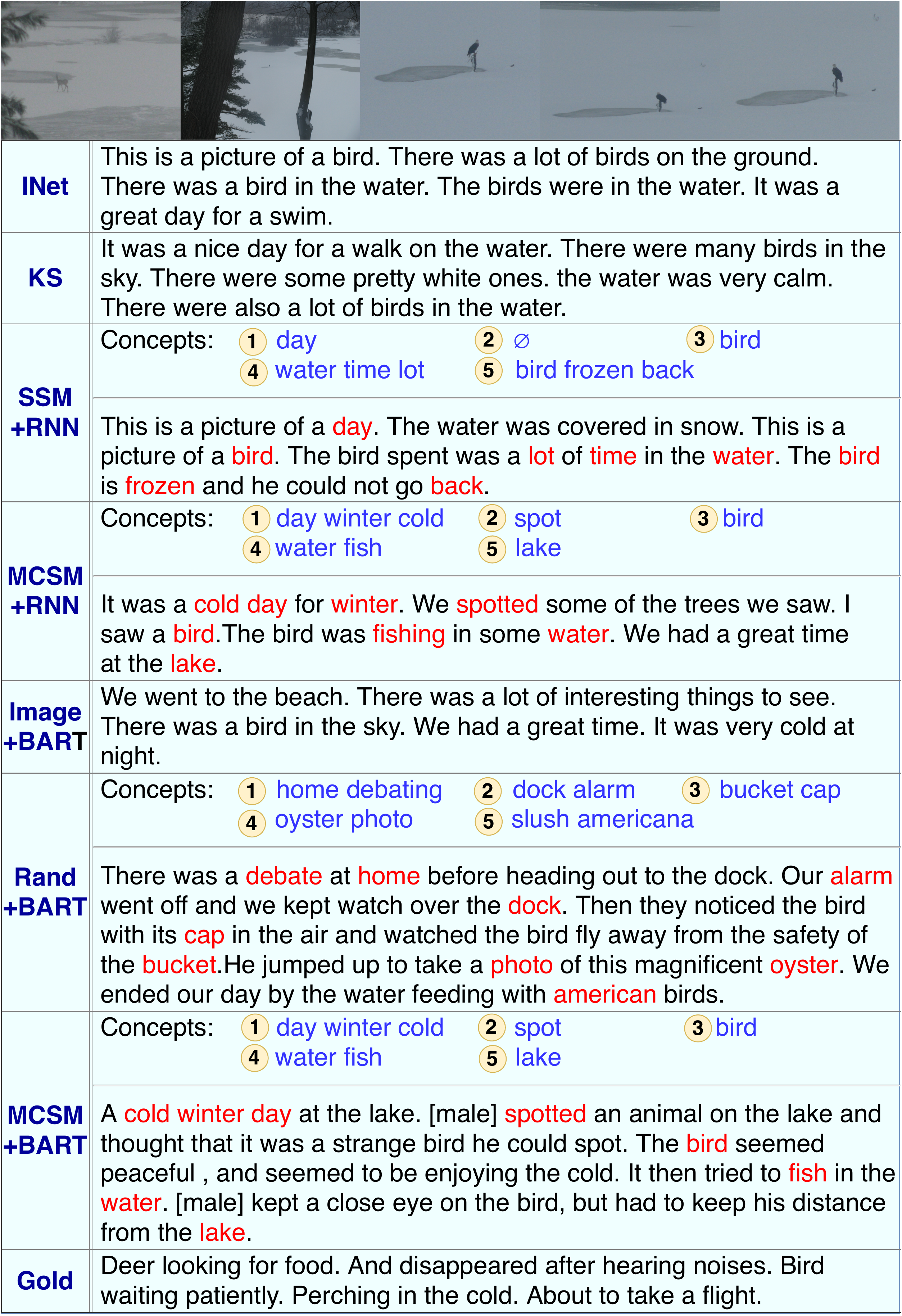}
\caption{The examples of generated stories by different methods. Our MCSM and SSM can generate better stories compared with other baselines that do not use BART. When using the pretrained BART, the concept selection with MCSM can produce vivid and informative story.}
\label{fig:demo}
\end{figure}

\subsection{Case Study}
We show a qualitative result of a random test sample in Figure~\ref{fig:demo}. This is a hard example because the last three images are very similar and the objects in all images are hard to recognize. We can see that INet generates monotonous and even irrelevant sentences. KS can generate better sentences but still low in lexical diversity. For the stories generated by two-stage strategy with RNN (SSM+RNN, MCSM+RNN), we can see that the story follows the selected concepts and the stories seem more reasonable than that of end-to-end training methods. 
When using BART, we compare three methods that represent no concept selection (Image+BART), bad concept selection (Rand+BART) and ours concept selection (MCSM+BART).
We can see that without using concepts or using randomly selected concepts, the generated stories are of low quality and to a certain extent irrelevant to the images. However, when guided by the selected concept, the story becomes vivid, relevant and logical.

\section{Conclusion}
In this work we exploit concept selection for improving the diversity and informativeness of stories generated from image sequences. By constructing a commonsense graph and two novel modules for concept selection, our proposed model outperforms all previous works in diversity by a large margin while still preserving the relevance and logical consistency on the VIST dataset. Our future direction aims to increase the relevance of the generated story by better leveraging the visual information.
% In this work, we aim to compose visual stories with high diversity and quality. Following plan-write strategy, we propose two novel concept selection models to plan concepts. Instead of generating the concepts directly from images, we construct a common sense graph which consists of candidate concepts. The experiment shows that both proposed selection module can achieve more accuracy than existing works. Afterwards, these concepts and images will be send into BART to generate the full story. The experiments show that our approach can outperform any existing works in diversity by a large margin while preserving the logical consistency and the relevance to the given images. Improving consistency and better use the information in visual information in large scale language model is our goal of this work.
% \chen{One of four parts need to be more emphasized: Concept selection, diversity, informativeness, BART.}
% \chen{$two stage generation \rightarrow concept selection, two stage generation \rightarrow diversity,
% commonsense knowledge \rightarrow informativeness, BART \rightarrow informativeness, nucleus sampling \rightarrow diversity, BART \rightarrow nucleus sampling, concept selection+BART \rightarrow remain Relevance$}

\section{Acknowledgements}
We thank the anonymous reviewers for the useful comments.
This paper was based on results obtained from a project, JPNP20006, commissioned by the New Energy and Industrial Technology Development Organization (NEDO) and was also supported by JSPS KAKENHI Grant Number JP19H04166.

\bibliography{cite}
\end{document}